\documentclass[bst/sn-basic]{sn-jnl}

\usepackage{graphicx}%
\usepackage{multirow}%
\usepackage{amsmath,amssymb,amsfonts}%
\usepackage[
        font={normal,small,sl,color=black!80},
        labelfont={normal,bf}
    ]{caption}
\usepackage{amsthm}%
\usepackage{mathrsfs}%
\usepackage[title]{appendix}%
\usepackage[]{mdframed}
\usepackage{xcolor}%
\usepackage{textcomp}%
\usepackage{manyfoot}%
\usepackage{booktabs}%
\usepackage{algorithm}%
\usepackage{algorithmicx}%
\usepackage{algpseudocode}%
\usepackage{listings}%
\usepackage{cite}
\usepackage{braket}
\usepackage{physics}
\definecolor{darkred}{rgb}{0.5,0.,0.}
\usepackage{orcidlink}

\usepackage{makecell}
\usepackage{natbib}



\begin{document}

\title[Article Title]{Hybrid Classical-Quantum Variational Autoencoder for Neural Topic Modeling}

\author*{\fnm{Ivan} \sur{Kankeu} \orcidlink{0009-0006-7470-4598}}\email{ikankeu.public@gmail.com}

\abstract{Neural topic models enable scalable semantic discovery, but their integration with quantum hardware remains largely unexplored. We present a proof-of-concept hybrid classical-quantum variational autoencoder (VAE) for topic modeling, embedding parameterized quantum circuits within the VAE inference network while retaining a classical topic-word decoder. To address the resource constraints of quantum hardware, we propose a modified Gaussian Softmax posterior that decouples latent space dimensionality from the number of topics to be extracted, enabling the model to operate with a low-resource 10-qubit quantum device. On the AgNews dataset, the hybrid VAE outperforms state-of-the-art neural topic models (NTMs), reaching a $C_v$ coherence score of 0.71 and an NPMI score of 0.20 while preserving high topic diversity. For comparison, we also construct a fully classical variant, which also outperforms state-of-the-art models on AgNews and exhibits clear class separation in the latent space. These results demonstrate that hybrid VAEs are computationally viable even on NISQ-era devices and represent a promising direction for quantum-enhanced topic modeling.}


\maketitle

\vspace{2.5cm}
\section{Introduction}\label{sec1}
Topic modeling is a machine learning technique used to unveil latent topics within a set of documents, generally in an unsupervised fashion. Topics are represented by sets of semantically related words that collectively describe coherent and distinct semantic concepts. For instance, the topic ``politics'' may be described by words such as ``law,'' ``policy,'' and ``government''. Owing to its interpretability, topic modeling has seen widespread use in applications such as text analysis, document retrieval, and content recommendation. Orthodox approaches to topic modeling include Bayesian probabilistic models, such as Latent Dirichlet Allocation (LDA), and matrix factorization methods. LDA \citep{a109}, one of the most popular techniques, uses Bayesian inference to discover latent topics by treating documents as distributions over topics. Matrix factorization, on the other hand, decomposes a word-document matrix into two lower-rank matrices: one describing word-topic relationships and the other describing topic-document relationships.

Conventional methods, however, face major challenges, including poor scalability to large datasets and an inability to capture non-linear relationships between topics and words. To overcome these limitations, Neural Topic Models (NTMs) have emerged as alternatives that can be trained efficiently and flexibly on large datasets using GPUs. In this paper, we propose and analyze the performance of a hybrid version of a popular NTM, namely the Variational Autoencoder (VAE).

Motivated by recent advances in quantum machine learning, this work explores how parameterized quantum circuits can be integrated into a VAE-based topic model while preserving competitive performance. Specifically, we embed quantum components inside the VAE inference network and compare the resulting hybrid models with fully classical counterparts on standard topic modeling benchmarks. Through this study, we provide a proof of concept for hybrid classical-quantum neural topic modeling and assess how the quantum component affects topic coherence, diversity, and latent space organization.

\section{Related Work}\label{sec2}

In contrast to a standard autoencoder, which offers a deterministic point estimate for the latent representation, a VAE produces latent variables that describe distributions over the latent space. While the latent space in a traditional autoencoder is typically sparse and disjointed, the VAE latent space is smooth and continuous, enabling meaningful and consistent sampled data points (topics). Therefore, a VAE is better equipped to extract and structure topics from a document collection. 

A VAE-based NTM comprises an inference network (encoder) and a generative network ($softmax$ decoder). The encoder infers latent variables from a bag-of-words (BoW) document representation $x$, parameterizing distributions from which the topic distribution vector $z$ is sampled. The assumption is that the document collection follows a prior distribution $p(z)$, which can be approximated to determine the topic distributions among documents. Thus, the encoder $\phi$ computes $q_\phi(z|x)$, a variational approximation to \(p(z|x) = \frac{p(z,x)}{p(x)}\), which is intractable because of the integral in \(p(x) = \int p(x|z)p(z)dz\). The approximation is obtained by minimizing the Kullback-Leibler (KL) divergence \(KL[q_\phi(z|x) || p(z|x)] = \log p(x) - ELBO\), where $ELBO$ is the Evidence Lower Bound. Since $\log p(x)$ is constant with respect to $z$, this is equivalent to minimizing \(-ELBO = KL[q_\phi(z|x) || p(z)] - \mathbb{E}_{q_\phi(z|x)}(\log p(x|z))\). $p(x|z)$, the conditional distribution, is provided by the decoder $\theta$, which reconstructs $x$, the original document representation. Hence, the loss function is expressed as:

\begin{equation}
\mathcal{L}_{\phi,\theta} = -ELBO = KL[q_\phi(z|x) || p(z)] - \mathbb{E}_{q_\phi(z|x)}(\log p_\theta(x|z)),
\end{equation}
where the second term is the reconstruction loss, and the first term, the KL divergence, regularizes the variational distribution to be close to the prior distribution. The decoder is typically a bias-free fully connected layer $W$ with \(p_\theta(x|z) = softmax(Wz)\).

Ordinarily, the prior distribution is assumed to be Gaussian, so the encoder infers the mean and the logarithm of the standard deviation of the distribution \((\mu,\log\sigma)\). Instead of \(z \sim \mathcal{N}(\mu,\sigma)\), the reparameterization trick is applied as \(z = \mu + \sigma\epsilon\) with \(\epsilon \in \mathcal{N}(\mathbf{0},\mathbf{I})\) to alleviate instability during training \citep{a112}. However, in topic modeling, the Gaussian prior is not always well-suited, as it tends to push the topic means in latent space toward the center, thereby entangling topics. Therefore, alternative approaches have been proposed to approximate the prior $p(z)$.

\cite{a111} introduced the Gaussian Softmax (GSM) technique, which applies a linear transformation to $z$ followed by a softmax activation function: \(g(z) = softmax(Wz + b)\). A key advantage of GSM is that the latent space dimension can be smaller than the topic count -- a feature we harness later. \cite{a110} propose approximating the Dirichlet multinomial distribution using a Laplace approximation. The Dirichlet distribution is particularly useful because it is defined over a $(K-1)$-dimensional simplex, allowing control over the distribution of topic proportions. \cite{a113} suggest leveraging semantically rich word embeddings to enhance topic models by factorizing the topic-word matrix into a product of topic embedding and word embedding matrices. Building upon previous work, \cite{a114} proposed vONTSS, an NTM based on optimal transport (also used in \citeyear{a216}) that outperforms existing NTMs in an unsupervised setting on the AgNews and 20News benchmark datasets. The vONTSS encoder $\phi$ maps a bag-of-words (BoW) document representation into \((\mu,k)\) parameters of the von Mises-Fisher (vMF) distribution, from which a latent vector $\eta$ is sampled. The vMF distribution is employed to mitigate the topic entanglement often ascribed to the Gaussian prior by restricting expressibility. Additionally, the vector $\eta$ is passed through a temperature function before the $softmax$ application to tune the resulting topic distribution vector $z$. From $z$, the decoder reconstructs the original BoW representation using a trainable topic embedding matrix and a frozen word embedding matrix.

Beyond classical autoencoders, quantum and hybrid autoencoders have gained significant attention in the quantum machine learning literature, offering potential advantages over their classical counterparts. Some of these approaches have already been experimentally implemented on real quantum devices, demonstrating their feasibility on near-term quantum hardware. In a pioneering study, \cite{a200} introduced a quantum autoencoder designed to compress quantum states beyond classical capabilities. Unlike classical autoencoders, where both input and output remain classical, their approach directly maps a quantum input to a quantum output, performing compression intrinsically within the quantum circuit. Another notable quantum generative model was introduced by \cite{a201}, who developed the first quantum VAE based on annealing-based generative models. This work later found applications in generative chemistry \citep{a207}, demonstrating the potential of quantum-enhanced VAEs for modeling complex probability distributions.  

More recently, hybrid quantum-classical autoencoders have emerged as promising architectures. These architectures integrate parameterized quantum circuits (PQCs) into classical autoencoder structures to improve various tasks, such as unsupervised dimensionality reduction and anomaly detection \citep{a209,a202}. A common approach involves using a classical encoder to generate a latent representation, which is then processed by a PQC before being measured, yielding expectation values that serve as inputs to a classical decoder. Alternatively, some approaches replace the classical encoder or decoder entirely with a PQC, allowing for direct extraction of information from quantum states or learning the probability distribution of quantum measurements \citep{a203,a204}. Notably, these methods have been shown to efficiently learn hard quantum states (e.g., Haar random states) with only a linear number of parameters, unlike classical models, which scale exponentially.

Several recent works adopt a NISQ-oriented strategy in which a classical model first compresses the input and the quantum component operates only on the resulting low-dimensional latent space. \cite{a210} trained a ResNet10-inspired convolutional autoencoder for image reconstruction, isolated its 64-dimensional latent representation, amplitude-encoded it into six qubits, and used QSVM/QOCSVM blocks for classification and anomaly detection. Their results show that the downstream quantum block can work well when the reconstruction latent space preserves discriminative information, but that performance degrades on more abstract or imbalanced image data. In a related MNIST study, \cite{a213} compressed images into 64 autoencoder features, further reduced them to five principal components, mapped them to a 5-qubit circuit, and classified the resulting 32-dimensional measurement distribution; the hybrid model remained functional but fell behind the classical latent space baseline, illustrating the cost of aggressive compression before quantum encoding. Outside image data, \cite{a212} combined classical autoencoders with quantum neural networks for heart disease classification and reported competitive accuracy under limited-data and noisy-simulation settings. These studies suggest that classical encoders are currently a practical bridge to quantum processing, while also making the information bottleneck a central design constraint.

To the best of our knowledge, our work is the first to apply a hybrid classical-quantum VAE with PQCs for topic modeling. Unlike the above pipelines, our PQCs are embedded inside the VAE inference network to parameterize the posterior distribution used to infer document-topic mixtures, while the decoder remains a topic-word reconstruction module tailored to NTM evaluation.

\section{Proposed Methods}\label{sec3}

Up to this point, we have provided a first glimpse of hybrid neural networks and neural topic modeling. As mentioned above, in topic modeling we have to deal with large datasets, which compels us to devise sophisticated techniques that aptly balance scalability and performance. In the following sections, we propose a novel technique based on VAEs and hybrid computing as a proof of concept for tackling topic modeling using quantum devices. Alongside the hybrid VAE, we also present a fully classical counterpart that is used to comparatively assess the performance of the proposed technique.

\begin{center}
   \includegraphics[width=13cm, height=4.5cm]{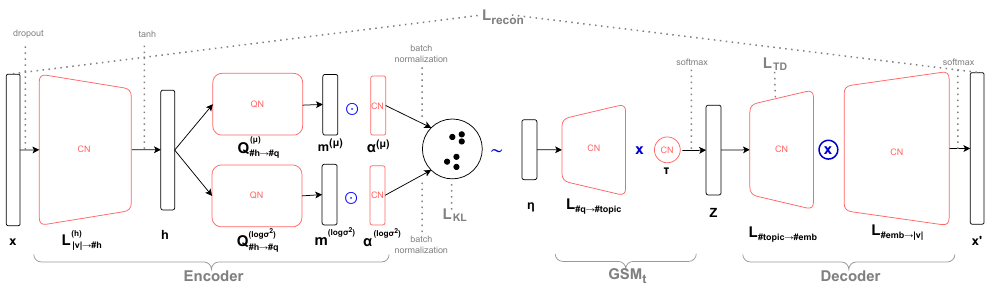}
   \captionof{figure}{Architecture of the hybrid VAE. The red components are the trainable nodes, where CN and QN stand for classical node and quantum node, respectively.}
    \label{fig3.5}
\end{center}

\subsection{Hybrid VAE}
Figure \ref{fig3.5} illustrates the model architecture at a high level. This model is designed to comply with VQA constraints. At an overarching level, the encoder network transforms a BoW document representation into the mean and logarithm of the variance of a Gaussian Softmax distribution, from which a topic distribution vector is sampled and decoded using a topic-word matrix to reconstruct the BoW representation.

The encoder network $\phi$ comprises a classical component and a quantum component. Starting with a BoW document representation $x\in\mathcal{R}^{|V|}$, the model down-projects it to a \#h-dimensional vector $h$ using a fully connected layer $\mathcal{L}_{|V|\to \#h}^{(h)}$, where $V$ is the vocabulary of terms appearing in the document collection. The hidden representation $h$ is then amplitude-encoded into VQCs (with padding), $\mathcal{Q}_{\#h\to\#q}^{(\mu)}$ and $\mathcal{Q}_{\#h\to\#q}^{(\log\sigma^2)}$, following Equation \ref{eq1.5} for $x\in\mathcal{R}^{N}$:
\begin{equation}\label{eq1.5}
    x \mapsto \frac{1}{||x||_2}\sum_{i=0}^{2^n-1} x_i\ket{i}.
\end{equation}
The VQCs $\mathcal{Q}_{\#h\to\#q}^{(\mu)}$ and $\mathcal{Q}_{\#h\to\#q}^{(\log\sigma^2)}$ compute the mean $\mu$ and the logarithm of the variance $\log\sigma^2$, respectively. They consist of sequences of strongly entangling layers, each parameterized with rotation gates (see Figure \ref{fig:apd.3}). Ultimately, we measure either the state probabilities of a subset of qubits or the Pauli Z expectation values for each qubit, producing \#q-dimensional classical vectors $m^{(\mu)}$ and $m^{(\log\sigma^2)}$. We restrict ourselves to these two types of measurements because both are amenable to differentiation methods, such as the parameter-shift rule and finite differences, using ``PennyLane'' \citep{d6}. Due to the high entanglement in the VQCs, the measurements need to be individually scaled by learnable \#q-dimensional vectors $\alpha^{(\mu)}$ and $\alpha^{(\log\sigma^2)}$ to increase the variance within the measurements -- an important step for alleviating topic entanglement and enhancing diversity. Consequently, we obtain \(\mu=m^{(\mu)}\odot\alpha^{(\mu)}\) and \(\sigma=\sqrt{e^{(m^{(\log\sigma^2)}\odot\alpha^{(\log\sigma^2)})}}\) as the mean and standard deviation of the Gaussian distribution over the latent space. Unlike typical approaches in the literature \citep{a111,a124}, which output $\log\sigma$, our method outputs $\log\sigma^2$, which slightly enhances performance. We apply a dropout layer to the input $x$ to prevent memorization, as well as two batch normalization layers to the scaled measurements to stabilize the training process.

The Gaussian Softmax distribution ($GSM_t$), parameterized by the encoder outputs, is modified and endowed with a learnable temperature parameter $\tau$, so that the topic distribution vector $z$ is sampled as follows: $z \sim G_{GSM_t}(\mu,\sigma^2)$. The sampling process utilizes the reparameterization trick:
\begin{equation}
\begin{split}
    \eta = \mu \oplus \epsilon\odot\sigma, \\
    z = softmax(\tau \mathcal{L}_{\#q\to\#topic}(\eta)), \\
    \text{with}\, \epsilon \sim \mathcal{N}(\mu,\sigma^2),
\end{split}
\end{equation}
where $\mathcal{L}_{\#q\to\#topic}$ is a fully connected layer without bias used to project into the topic vector space, whose dimension $\#topic$ is the number of topics to be extracted. The $GSM_t$ is aptly chosen to decouple the topic count from the encoder output dimension, which eases the integration of the quantum components. The temperature, on the other hand, is introduced to foster topic disentanglement.

For the decoder network, we follow the proposal of \cite{a113} and utilize topic embedding $\mathcal{L}_{\#topic\to\#emb}$ and word embedding $\mathcal{L}_{\#emb\to|V|}$ matrices to construct the topic-word matrix. Thus, $x'$ is reconstructed as follows:
\begin{equation}
x' = softmax(\mathcal{L}_{\#emb\to|V|}(\mathcal{L}_{\#topic\to\#emb}(z))),
\end{equation}
where $\mathcal{L}_{\#topic\to\#emb}$ and $\mathcal{L}_{\#emb\to|V|}$ are both fully connected layers without bias. In our experiments, we initialize the word embedding layer with pre-trained 300-dimensional GloVe embeddings ($\#emb=300$) \citep{a125} to improve training.

A topic diversity regularization term $\mathcal{L}_{TD}$ is introduced to foster diversity, following the recommendations of \cite{a111}. However, instead of using the $arccos$ function to capture angular distances as they did, we utilize cosine similarity. The purpose of this regularization is to encourage the model to learn orthogonal topic embedding vectors, thereby reducing topic entanglement. Hence, the regularization is calculated as the sum of the mean and variance of the absolute values of the topic embedding distances:
\begin{equation}
\begin{split}
    &\mathcal{L}_{TD} = \zeta + variance(\zeta), \\
    &\text{with}\, \zeta = \frac{1}{\#topic^2}\sum_{i=1}^{\#topic}\sum_{j=1}^{\#topic} |cos\_sim(t_i,t_j)|,
\end{split}
\end{equation}
where $t_i,t_j \in \mathcal{L}_{\#topic\to\#emb}$ are topic embeddings of length $\#emb$. To minimize the loss, the model must reduce both \(\zeta\) and \(\text{variance}(\zeta)\) towards 0 to promote orthogonality.

The final loss function $\mathcal{L}$ combines the reconstruction loss $\mathcal{L}_{recon}$, the topic diversity regularization $\mathcal{L}_{TD}$, and the Kullback-Leibler divergence $\mathcal{L}_{KL}$ between the posterior distribution and the standard Gaussian distribution. Ergo, we have
\begin{equation}
\begin{split}
    &\mathcal{L} = \mathcal{L}_{recon} + \mathcal{L}_{KL} + \mathcal{L}_{TD}, \\
    &\text{with}\, \mathcal{L}_{recon} = \sum_{i=1}^{|V|} x_i\log x_i', \, \mathcal{L}_{KL} = \frac{1}{2}\sum_{i=1}^{\#q}(-1 -2\log\sigma_i + \sigma_i^2 + \mu_i^2), 
\end{split}
\end{equation}
where $\mu_i$ and $\sigma_i$ are the i-th elements of the vectors $\mu$ and $\sigma$, respectively.

\subsubsection{Quantum components}
Although the functional form of the encoder has been defined, the specific quantum components have not yet been specified. As illustrated in Figure \ref{fig3.5}, the encoder \(\phi\) consists of two PQCs following a fully connected down-projection layer. The hidden representation $h$ is amplitude-encoded into a quantum state as
\begin{equation}
\ket{\psi_h} = S(h) \ket{0}^{\otimes \#q} = \frac{1}{\|h\|} \sum_{i=0}^{2^{\#q}-1} h_i \ket{i},
\end{equation}
where $S(h)$ represents an arbitrary state preparation routine parameterized by angles derived from $h$. The PQCs then process the quantum state $\ket{\psi_h}$, applying the unitary transformations $U(\theta^{(\mu)})$ and $U(\theta^{(\log\sigma^2)})$, corresponding to the mean and logarithm of the variance, respectively. 

To extract classical information, the expectation value of the Pauli-Z operator on each qubit is measured. These expectation values, combined with trainable $\alpha$-vectors, are used to compute the parameters of the distribution. Specifically, the classical vectors obtained from the quantum circuits are given by
\begin{equation}
m^{(\mu)}_i = \bra{\psi_h} Z_i U(\theta^{(\mu)}) \ket{\psi_h}, \quad 
m^{(\log\sigma^2)}_i = \bra{\psi_h} Z_i U(\theta^{(\log\sigma^2)}) \ket{\psi_h},
\end{equation}
where $Z_i$ denotes the Pauli-Z operator acting on the i-th qubit. Thus, the complete dressed quantum circuit for the encoder can be expressed as
\begin{equation}
\phi = \mathcal{L}_{|V| \to \#h}^{(h)} \circ \mathcal{Q}_{\#h \to \#q}^{(*)} \circ \alpha^{(*)},
\end{equation}
which maps the BoW representation to the mean and logarithm of the variance of the distribution.

\subsection{Classical VAE}
To construct the classical counterpart, we replace both VQCs with fully connected layers followed by $tanh$ activation functions. Additionally, we remove the temperature parameter as well as the parameter vectors $m^{(\mu)}$ and $m^{(\log\sigma^2)}$.

\section{Experiments}\label{sec4}

Given the significant impact of dataset pre-processing on evaluation results \citeyearpar{a126,a127,a128}, we strictly adhere to the pre-processing steps outlined in the paper that reports state-of-the-art (SOTA) performance \citep{a114}. To this end, we reuse their publicly accessible pre-processing algorithm \citeyearpar{c1} without modifications. This allows us to compare their evaluation results effectively with our own. We test both hybrid and classical models in two variants: a small latent space (SLS) with $\#q = 10$ and a large latent space (LLS) with $\#q = 32$. For the classical model, we simply adjust the size of the fully connected layers accordingly. For the hybrid model, we use 10 qubits and adjust the measurement method, using the Pauli Z expectation values with respect to every qubit for $\#q = 10$ and the state probabilities of the last 5 qubits for $\#q = 32$. Thus, we have two hybrid models and two classical models. Herein, we detail the datasets, evaluation metrics, and experimental settings used in our study.

\paragraph{Task:} The objective is to extract 20 topics from each of the benchmark datasets by training the models in an unsupervised manner. The decoder network of the model provides the solution by encoding the probability distribution over words for every topic. Thus, both the hybrid and classical model architectures are adjusted to feature decoders that resemble $20\times|V|$ matrices.

\paragraph{Datasets:} For training and evaluation, we utilize the following popular benchmark datasets:
\begin{enumerate}
    \item[-]{20News \citep{a129}}: This dataset consists of approximately 18,000 newsgroup posts across 20 topics, including religion, politics, and sports. It can be accessed using the ``octis'' library. During pre-processing, each document was tokenized and cleaned, with stop words and words occurring in more than 15\% of all documents or fewer than 20 times removed \citeyearpar{a114}.
    
    \item[-]{AgNews \citep{a130}}: The AG's News dataset comprises four classes, each containing 30,000 news articles sourced from the web and covering topics such as sports and business. It can be retrieved using the ``Hugging Face'' datasets library. Its pre-processing is the same as that of the 20News dataset.
\end{enumerate}

\vspace{.7cm}
\begin{center}
\captionof{table}{Statistics of the topic modeling benchmark datasets used}
\begin{tabular}{c c c}
 \toprule
 \textbf{Datasets} & \textbf{\#Documents} & \textbf{\#Terms} \\
 \midrule
 20News & 16,309 & 1,369 \\ 
 \hline
 AgNews & 120,000 & 14,696 \\ 
 \hline
\end{tabular}
\label{tab3.1}
\end{center}

\paragraph{Evaluation Metrics:} To assess the quality of the learned topic-word matrix, we employ standard metrics widely used in topic modeling:
\begin{enumerate}
    \item[-]{NPMI (Normalized Pointwise Mutual Information) \citeyearpar{a132}} gauges the strength of the relationship between word pairs within topics based on their co-occurrence in a document collection. It is defined as \(NPMI(w_i,w_j) = \frac{PMI(w_i,w_j)}{\log P(w_i,w_j)}\), where \(PMI(w_i,w_j) = \log\frac{P(w_i,w_j)}{P(w_i)P(w_j)}\) is the pointwise mutual information, $P(w_i)$ is the probability of word $w_i$ occurring, and \(P(w_i,w_j)\) is the probability of both words co-occurring.
    \item[-]{$C_v$ Coherence \citeyearpar{a131}} evaluates topic coherence based on NPMI and word co-occurrence. It posits that words within the same topic should frequently co-occur in the document collection.
    \item[-]{TD (Topic Diversity) \citeyearpar{a113}} measures the diversity of the discovered topics. It is defined as the ratio of unique words among the top-K words across all topics.
    \item[-]{Quality \citeyearpar{a113}} is a derived metric introduced by \citeauthor{a113} to provide a single score reflecting topic quality. It is calculated as the product of coherence and topic diversity. In our experiments, we use $C_v$ for coherence, as it offers a more robust evaluation than NPMI.
\end{enumerate}
Higher scores for these metrics are better, with 1 indicating the best possible score. In our implementation, we use the top 10 topic words to compute $C_v$ and NPMI, while the top 25 topic words are used for TD, following the settings in the paper with SOTA performance \citeyearpar{a114}. We employ the Python library ``octis'' \citeyearpar{a128} to calculate all the aforementioned metrics.

\paragraph{Training settings:} For our experiments, we set the learning rate to 2e-3 and the batch size to 200, employ the Adam optimizer, and train for 20 epochs. Each model is trained five times, with the seed values for the ``torch(cuda)'', ``numpy'', and ``math'' libraries incremented by 1, starting from 42 and ending at 46. This approach ensures consistent initialization conditions, including the same training set and initial parameters. We use a single GPU as hardware. The quantum device is simulated without noise using PennyLane's ``default.qubit'' device, with gradients computed via backpropagation. Nevertheless, our hybrid model is fully compatible with differentiation methods such as parameter-shift and finite differences, allowing the quantum components to be executed on either simulated noisy quantum devices or actual quantum hardware.

\paragraph{Testing settings:} After each epoch, we assess the performance of the trained model on the benchmark datasets 20News and AgNews using the metrics $C_v$, NPMI, and TD. The 20 topics are derived from the decoder network by applying a $softmax$ function to the topic-word matrix and selecting the top-K words for each topic dimension. Additionally, we save the log data generated during training and evaluation in a ".txt" file. The log file follows this format:
\vspace{.25cm}
\begin{mdframed}
\begin{small}
\begin{verbatim}
Start Training: [date time]
Model: _, Pretrained Model: _
Dataset: [dataset name], Batch: [size], GPU: [ID]
Epoch _, Train Loss: [training loss]
Epoch _, CV: _, NPMI: _, TD: _
End Training: [date time]
\end{verbatim}
\end{small}
\end{mdframed}
\vspace{.25cm}
Any information not specified in this format but included in the log files is considered irrelevant and can be disregarded.

\section{Result Analysis \& Discussion}\label{sec5}

\begin{figure}[!htb]
   \begin{minipage}{0.49\textwidth}
        \centering
        \includegraphics[height=4cm]{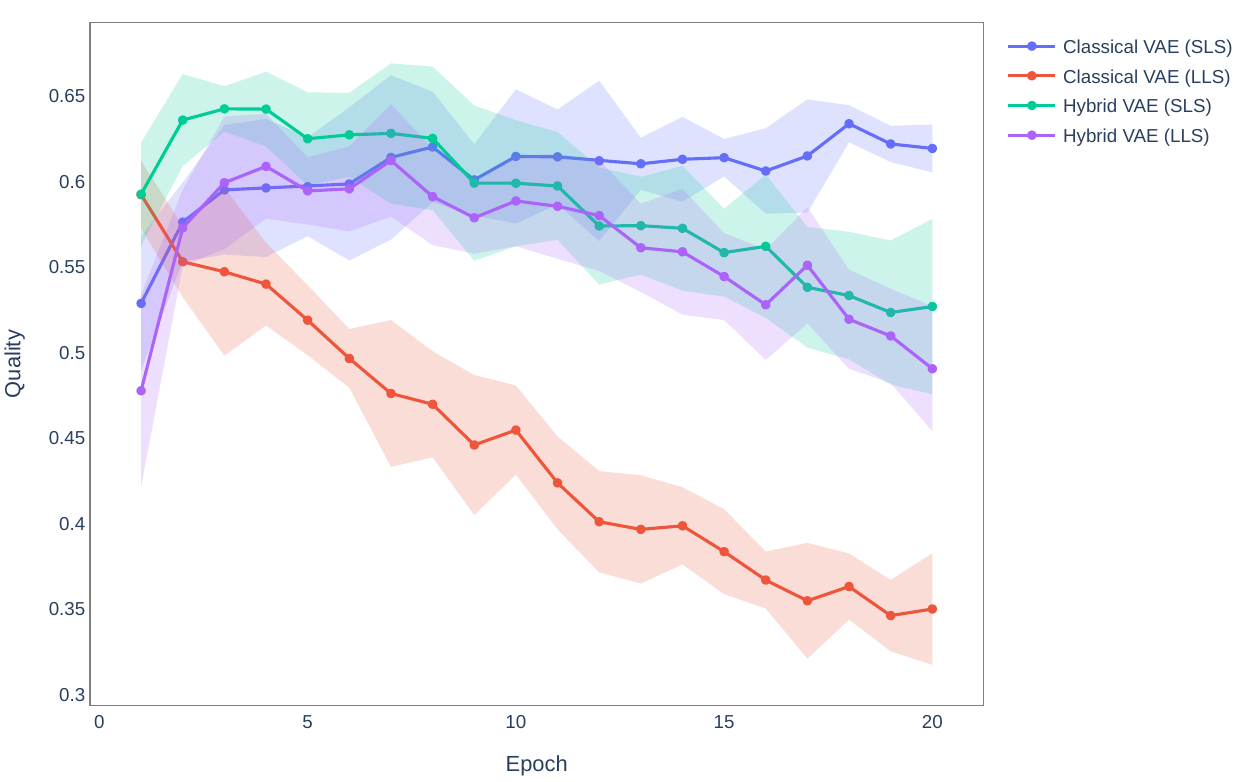}
        \captionof{figure}{Averages and standard deviations of model quality scores across 5 runs on AgNews during training.}
        \label{fig3.6}
   \end{minipage}\hfill
   \begin{minipage}{0.49\textwidth}
        \centering
        \includegraphics[height=4cm]{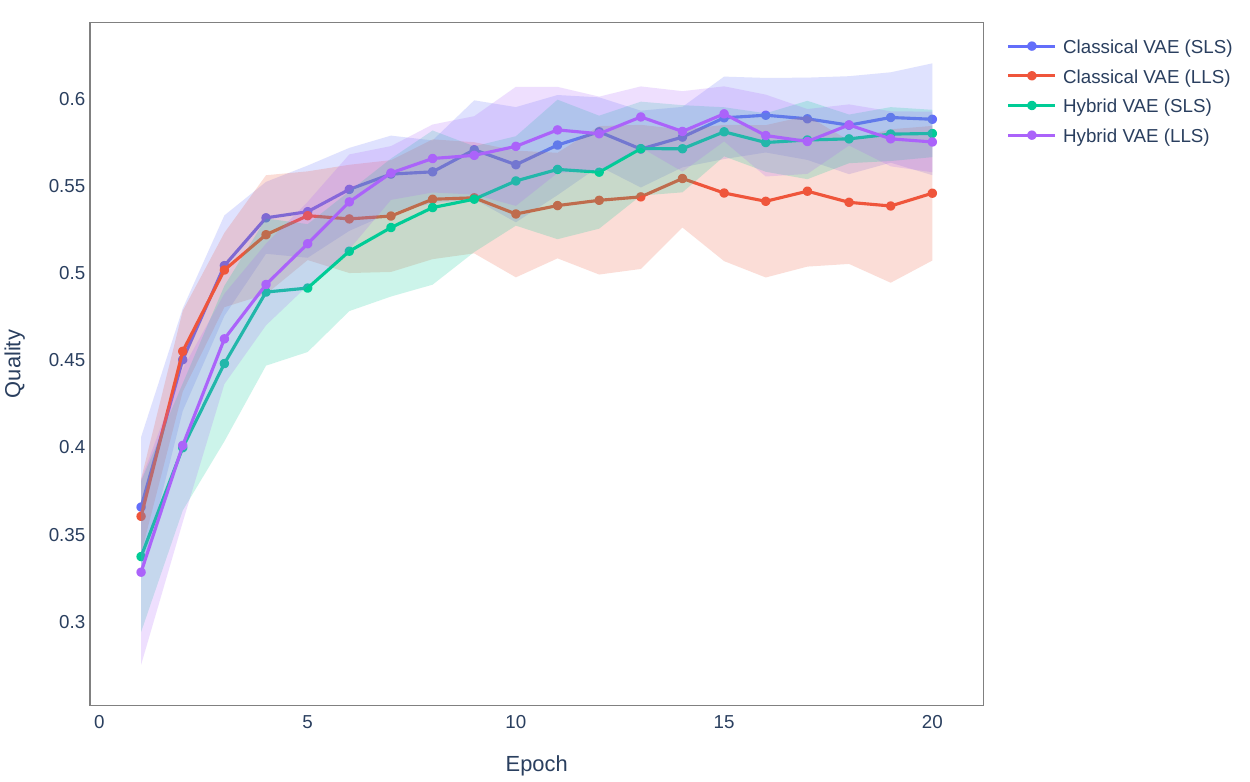}
        \captionof{figure}{Averages and standard deviations of model quality scores across 5 runs on 20News during training.}
        \label{fig3.7}
   \end{minipage}
\end{figure}

So far, we have outlined the model architectures and experimental settings. In this section, we analyze and discuss the results from our experiments for the four different models: classical VAE (SLS), classical VAE (LLS), hybrid VAE (SLS), and hybrid VAE (LLS). Figures \ref{fig3.6} and \ref{fig3.7} report the quality scores aggregated over five independent runs, providing a basis for comparing the models' performance. Figure \ref{fig3.10} illustrates the latent spaces of the trained models.

We begin by examining the training dynamics of the models. Figures~\ref{fig3.6} and~\ref{fig3.7} show the evolution of aggregated quality scores over 20~epochs for AgNews and 20News. Two consistent trends emerge. First, the hybrid models exhibit similar performance regardless of the latent space configuration, with curves closely following each other. Second, while models trained on 20News gradually seem to converge toward an upper plateau, those trained on AgNews reach peak performance rapidly before their scores begin to deteriorate.

The first trend indicates that the choice of latent space configuration (SLS vs.\ LLS) has limited influence on the performance of the hybrid models. Since the two configurations differ only in their measurement strategy, the resulting representations passed from the quantum component to the classical component likely remain similar in informational content. This stands in contrast to the classical models, where the LLS configuration employs a larger fully connected layer with more parameters, producing output vectors that are not only higher-dimensional but also richer in information.

The second trend can be attributed to the disparity in dataset sizes: AgNews contains nearly seven times more training samples than 20News. This abundance allows models trained on AgNews to reach their optimal performance quickly, after which overfitting causes a decline. In contrast, models trained on 20News start with lower scores and improve gradually, seeming to progress toward a plateau. A closer inspection (Figures \ref{fig:apd.4}--\ref{fig:apd.7}) reveals that the decline in quality scores is primarily driven by decreasing $C_v$ coherence scores, whereas topic diversity increases and remains relatively stable. This pattern suggests that, as training progresses, the models tend to generate less coherent topics, a typical symptom of overfitting. Notably, this issue appears less severe in hybrid models and is exacerbated by the latent space configuration in classical models. 

Table \ref{tab3.2} compares our models with the previously reported SOTA results \citeyearpar{a114}. The reported SOTA scores are averaged over ten runs, whereas our results are averaged over five runs.

\vspace{.25cm}
\begin{tiny}
\begin{center}
\bgroup
\def\arraystretch{1.5}%
\captionof{table}{Comparison of model performance. Our results are averaged over five runs. * indicates results reported from the paper presenting the SOTA performance \citeyearpar{a114}.}
\begin{tabular}{c | c c c | c c c}
 \toprule
 \multirow{2}{*}{\textbf{Models}} & \multicolumn{3}{c|}{\textbf{AgNews}} & \multicolumn{3}{c}{\textbf{20News}} \\
 & \textbf{$C_v$}  & \textbf{NPMI} & \textbf{TD}  & \textbf{$C_v$}  & \textbf{NPMI} & \textbf{TD} \\
 \midrule
GSM* & 0.41$\pm$0.01 & 0.03$\pm$0.01 & 0.58$\pm$0.02    &    0.55$\pm$0.04 & 0.07$\pm$0.03 & 0.66$\pm$0.05\\
\hline
vONT* (SOTA) & 0.49$\pm$0.02 & 0.054$\pm$0.02 & 0.99$\pm$0.01    &    \textbf{0.69$\pm$0.03} & \textbf{0.16$\pm$0.02} & 0.96$\pm$0.03\\
\hline
& \multicolumn{6}{|c}{Our models}\\
\hline
Classical VAE (SLS) & \textbf{0.7$\pm$0.02} & \textbf{0.19$\pm$0.02} & 0.93$\pm$0.02     &   0.72$\pm$0.02 & 0.15$\pm$0.01 & 0.84$\pm$0.01\\
\hline
Classical VAE (LLS) & 0.65$\pm$0.02 & 0.15$\pm$0.01 & 0.92$\pm$0.03    &    0.69$\pm$0.03 & 0.14$\pm$0.02 & 0.82$\pm$0.02\\
\hline
Hybrid VAE (SLS) & \textbf{0.71$\pm$0.02} & \textbf{0.2$\pm$0.01} & 0.95$\pm$0.0    &    0.71$\pm$0.02 & 0.15$\pm$0.01 & 0.83$\pm$0.02\\
\hline
Hybrid VAE (LLS) & 0.65$\pm$0.04 & 0.16$\pm$0.03 & 0.96$\pm$0.01   &    0.73$\pm$0.03 & 0.16$\pm$0.01 & 0.82$\pm$0.01\\
\hline
\end{tabular}
\label{tab3.2}
\egroup
\end{center}
\end{tiny}

\paragraph{Performance on AgNews}

All of our models substantially outperform the SOTA in topic coherence on AgNews. The best-performing model, Hybrid VAE (SLS), achieves a ($C_v$) score of (0.71) and an NPMI of (0.20), compared to (0.49) and (0.054) for the SOTA model. Topic diversity is slightly lower than the SOTA but remains high across all variants.

The topic word clouds in Figure \ref{fig:apd.8} show that all models consistently identify the dominant topic keywords. Differences appear mainly in the probability distributions assigned to these words rather than in the discovered topics themselves.

Figure \ref{fig3.10} reveals an interesting observation: latent space separability does not correlate strongly with topic-modeling performance. The figure shows 1,000 randomly sampled latent vectors from three AgNews classes. The classical VAE (LLS) exhibits the clearest inter‑class separation, outperforming the latent space organization reported for the vMF‑based SOTA model. Yet this geometric clarity fails to translate into higher coherence or diversity scores. This disconnect suggests that current evaluation metrics capture latent space disentanglement only superficially. In particular, the topology of the learned latent space appears to correlate only weakly with the final performance metrics.

For the classical models, the higher-dimensional latent space appears to facilitate topic disentanglement and class separation. In contrast, the hybrid models do not benefit from increased latent dimensionality to the same extent. Despite operating in a high-dimensional latent space, the outputs of the quantum components remain relatively similar across samples, limiting separability.

Nevertheless, the hybrid models still achieve reasonable class separation and maintain competitive topic quality. Topic entanglement is partially mitigated by the learnable scaling vectors applied after the quantum circuit outputs. These scaling parameters increase output variance and improve class separation.

\begin{center}
   \includegraphics[width=12cm, height=6cm]{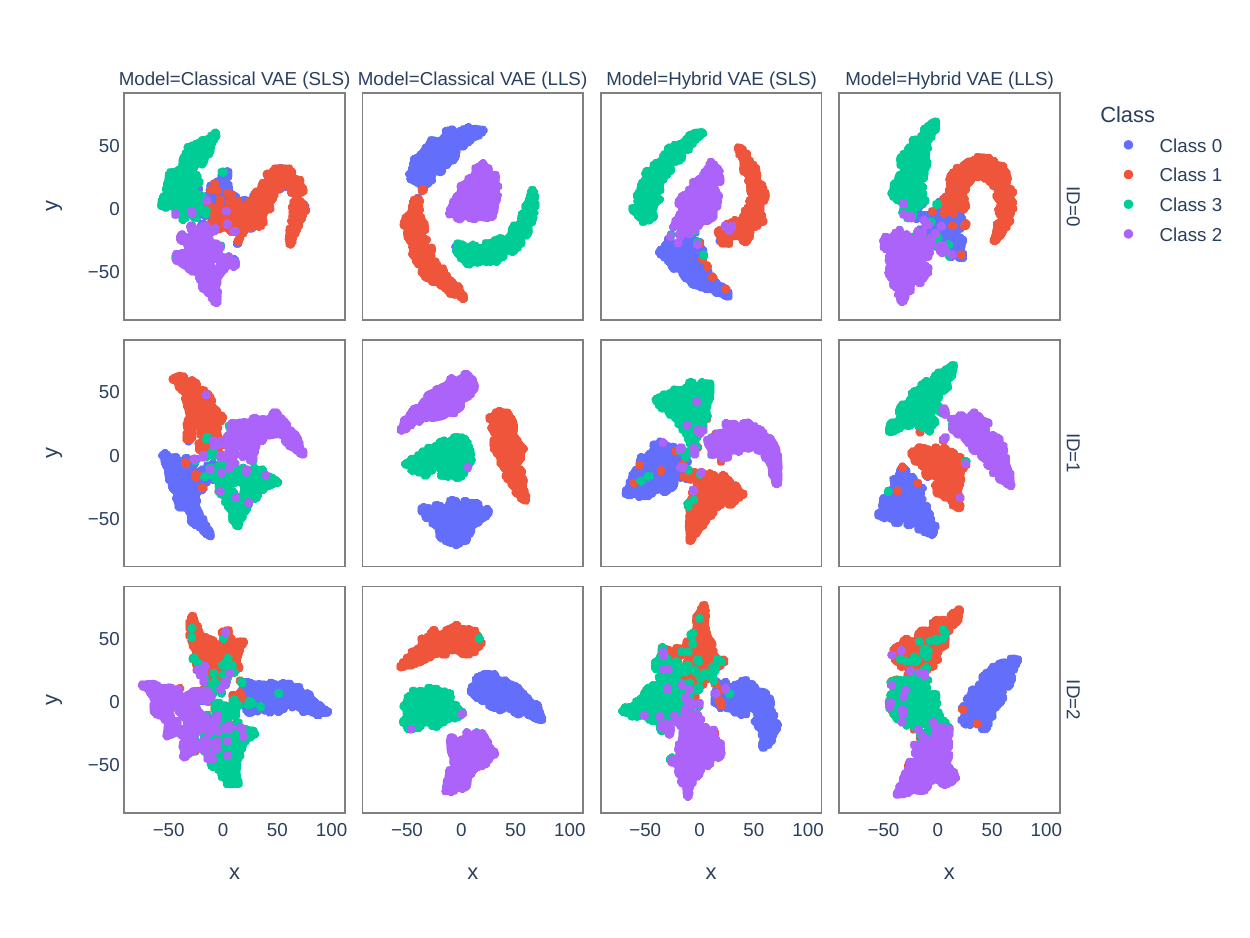}
   \captionof{figure}{2-D t-SNE projection of randomly sampled $z$ from latent spaces under 4 different posterior distributions on AgNews.}
    \label{fig3.10}
\end{center}

\paragraph{Performance on 20News}

On 20News, our models achieve coherence scores comparable to or slightly better than the SOTA. However, this improvement comes at the cost of reduced topic diversity.

Inspection of the topic word clouds (Figure \ref{fig:apd.10}) indicates that some topics are not captured consistently. For example, the medical topic generated by the Hybrid VAE (SLS) includes unrelated terms such as ``bike'' and ``motorcycle'', indicating residual topic mixing.

The latent space visualizations in Figure \ref{fig:apd.9} mirror the pattern observed on AgNews. The classical VAE (LLS) displays the most pronounced cluster separation, whereas the classical VAE (SLS) and the hybrid variants show noticeably weaker structure, with topics entangled near the center.

\paragraph{Discussion}

The results support two main conclusions.

First, hybrid VAEs are viable and trainable for neural topic modeling. We observe no severe optimization difficulties, significant performance degradation, or evidence of barren plateau effects \citeyearpar{a145}.

Second, the Gaussian Softmax (GSM) posterior enables effective decoupling between the latent representation and the quantum circuit architecture. This flexibility allows hybrid models to achieve performance comparable to purely classical alternatives.

However, the hybrid models still struggle to disentangle topics as effectively as the best classical variants. This limitation is partially alleviated through post-processing layers applied after the quantum circuit outputs. In further experiments (not reported here), post-processing fully connected layers did not produce better results than simple learnable scaling vectors  $m^{(\mu)}$ and $m^{(\log\sigma^2)}$, suggesting that the model can optimize a plain parameter vector more straightforwardly.

The GSM baseline, which also employs the Gaussian Softmax distribution, underperforms considerably relative to our models, likely due to inappropriate activation functions ($softplus$ instead of $tanh$), as well as the absence of batch normalization.

Finally, although not included in the reported experiments, we evaluated architectures with between zero (no trainable layer) and five trainable post-processing layers. None of these configurations produced a significant improvement over the reported results.

\section{Conclusion}\label{sec6}

In summary, we have successfully integrated a quantum component into the classical framework of a variational autoencoder. This was achieved by designing a VQC-friendly variational autoencoder that uses amplitude encoding and an enhanced Gaussian Softmax distribution. We minimized the resource requirements for the quantum component to ensure NISQ compatibility. Consequently, a quantum hardware setup with at least 10 qubits and mild error mitigation can effectively run the quantum circuit instead of relying solely on a quantum simulator. The measurement outcomes, whether 10 or 32 elements, provide an advantage in execution time and enable precise gradient calculation methods, such as the parameter-shift rule, which is essential for good approximations on NISQ hardware, even in the presence of noise. Therefore, aside from the extensive runtime required for training on large datasets, our hybrid VAE could be executed on quantum hardware without significant performance loss.

Moreover, while developing our hybrid model, we arrived at an architecture that surpasses SOTA results on AgNews by \textbf{45\%}, in both classical and hybrid implementations. This also shows that latent space dimensionality can be detached from the number of topics to be extracted, although this noticeably affects class separability. However, we emphasize that the primary aim of a hybrid model is not to surpass classical models, which can always be improved, but rather to serve as a proof of concept for hybrid computation with variational autoencoders that ideally preserves performance.

Future work could involve training our hybrid VAE on a small dataset using actual quantum hardware and conducting an in-depth study of the effects of noise on model trainability.

\section*{Data and Code Availability}
The source code used in this manuscript is available at \href{https://github.com/ikankeu-public/hcqvae}{https://github.com/ikankeu-public/hcqvae}.

\section*{Declarations}

\paragraph{Conflict of Interest} The authors declare no competing interests.

\paragraph{Generative AI} AI tools such as ChatGPT-4o and ChatGPT-5.5 have been used for grammar and
spelling checks, rephrasing, and rewording to improve the writing style. The generated text has been carefully read and verified to be free of hallucinations.

\bibliographystyle{bst/sn-basic}

\begin{appendices}

\section{Experiment Details}\label{secA4}

\subsection{Quantum circuit}

\begin{center}
   \includegraphics[width=13cm, height=3cm]{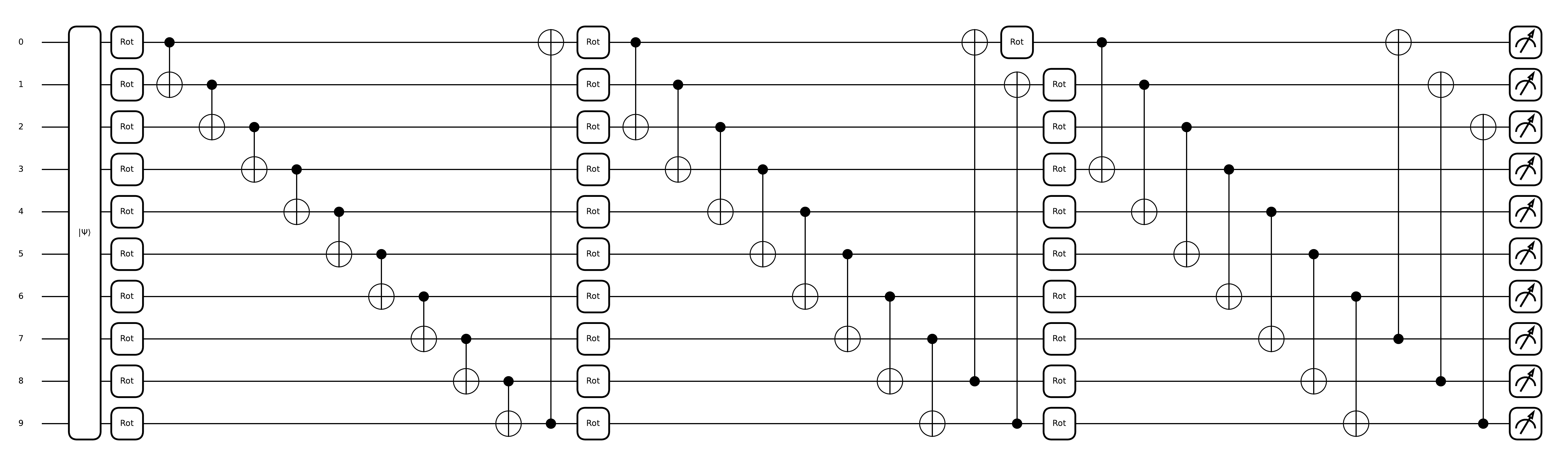}
   \captionof{figure}{Quantum circuit of the hybrid VAE.}
    \label{fig:apd.3}
\end{center}

\subsection{Result details}

\begin{figure}[!htb]
   \begin{minipage}{0.49\textwidth}
        \centering
        \includegraphics[height=4cm]{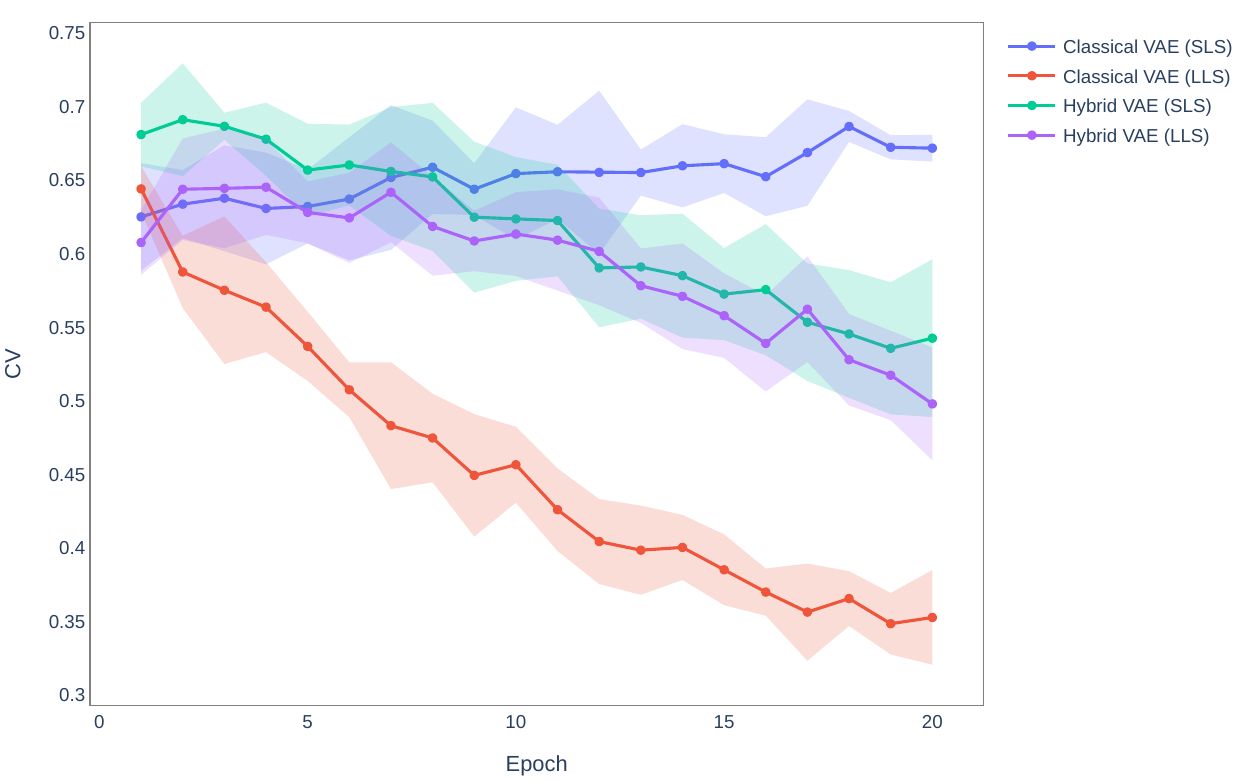}
        \captionof{figure}{Averages and standard deviations of model $C_v$ scores across 5 runs on AgNews during training.}
        \label{fig:apd.4}
   \end{minipage}\hfill
   \begin{minipage}{0.49\textwidth}
        \centering
        \includegraphics[height=4cm]{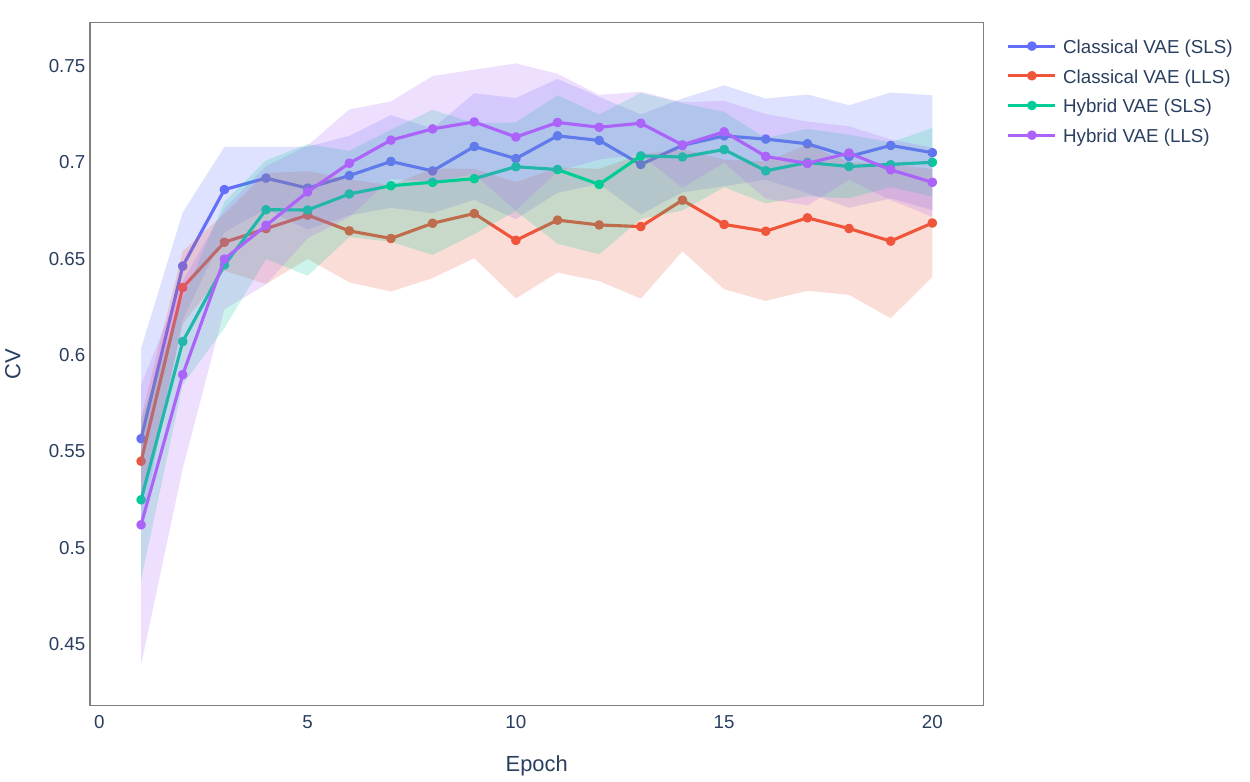}
        \captionof{figure}{Averages and standard deviations of model $C_v$ scores across 5 runs on 20News during training.}
        \label{fig:apd.5}
   \end{minipage}
\end{figure}

\begin{figure}[!htb]
   \begin{minipage}{0.49\textwidth}
        \centering
        \includegraphics[height=4cm]{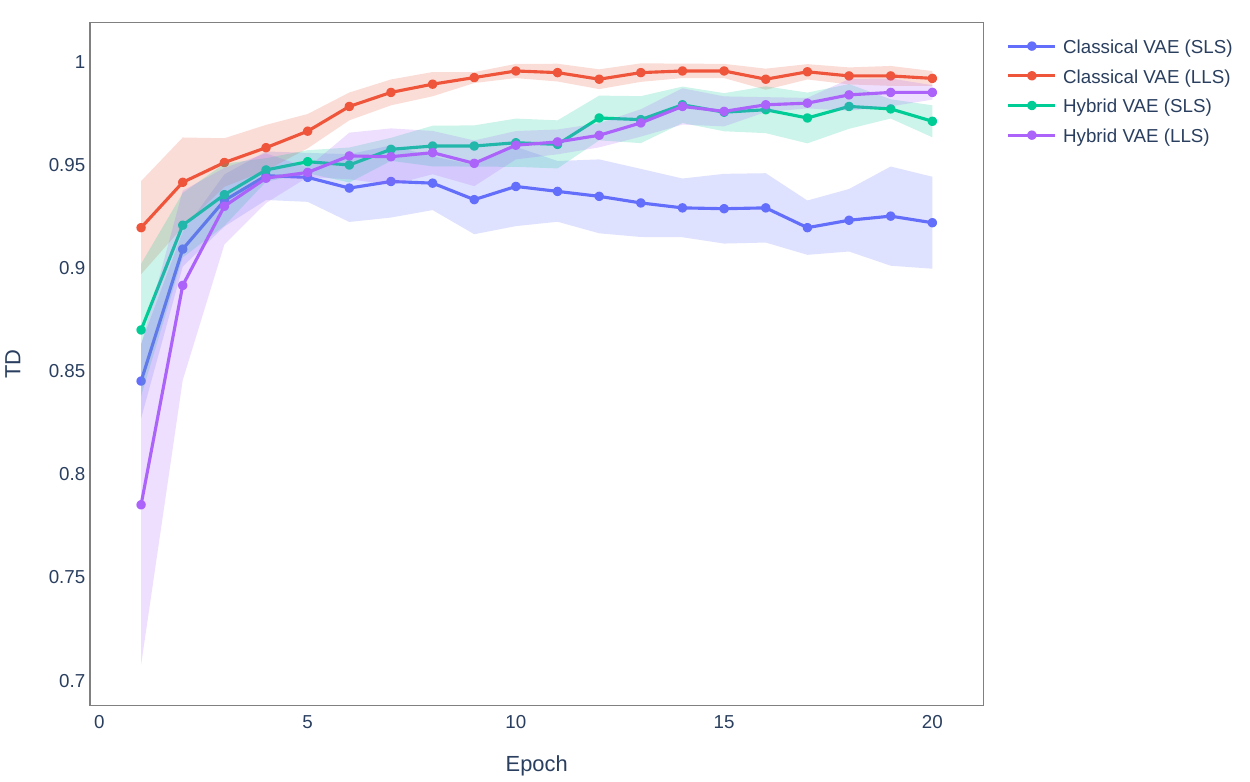}
        \captionof{figure}{Averages and standard deviations of model TD scores across 5 runs on AgNews during training.}
        \label{fig:apd.6}
   \end{minipage}\hfill
   \begin{minipage}{0.49\textwidth}
        \centering
        \includegraphics[height=4cm]{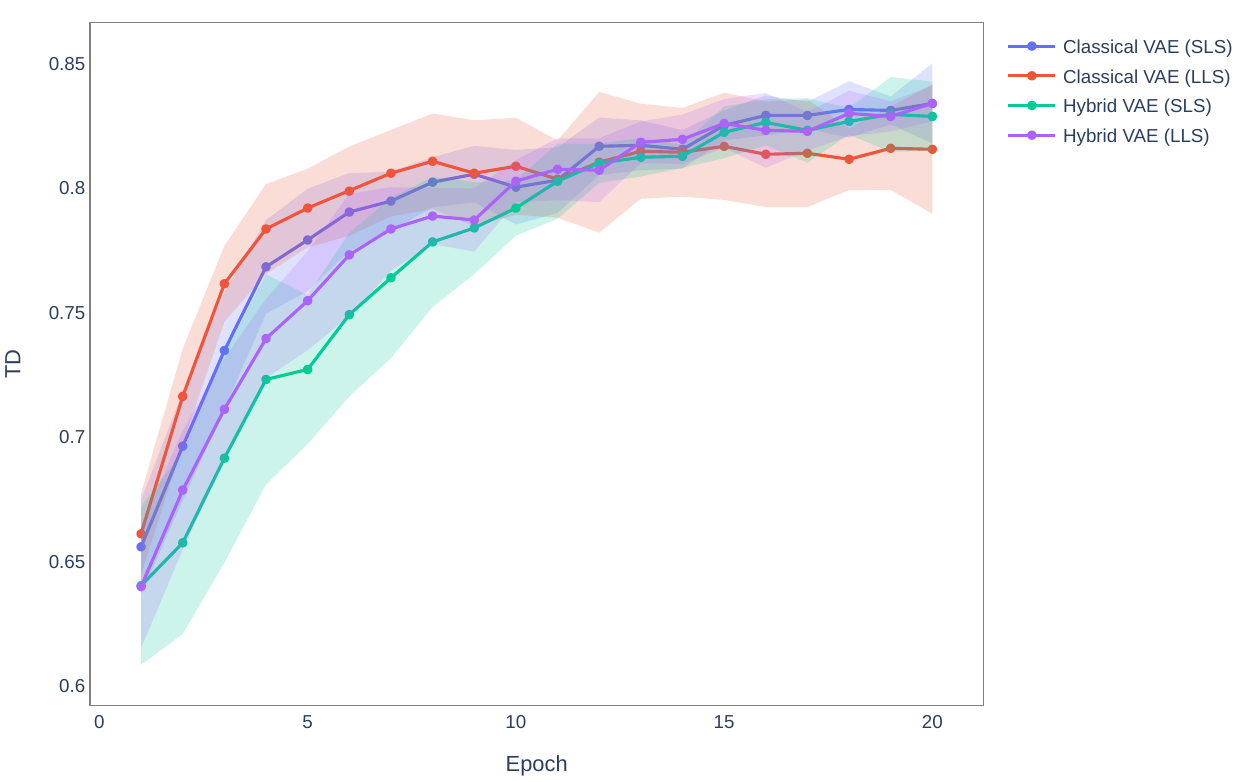}
        \captionof{figure}{Averages and standard deviations of model TD scores across 5 runs on 20News during training.}
        \label{fig:apd.7}
   \end{minipage}
\end{figure}

\vspace{.25cm}
\begin{center}
   \includegraphics[width=13cm, height=8cm]{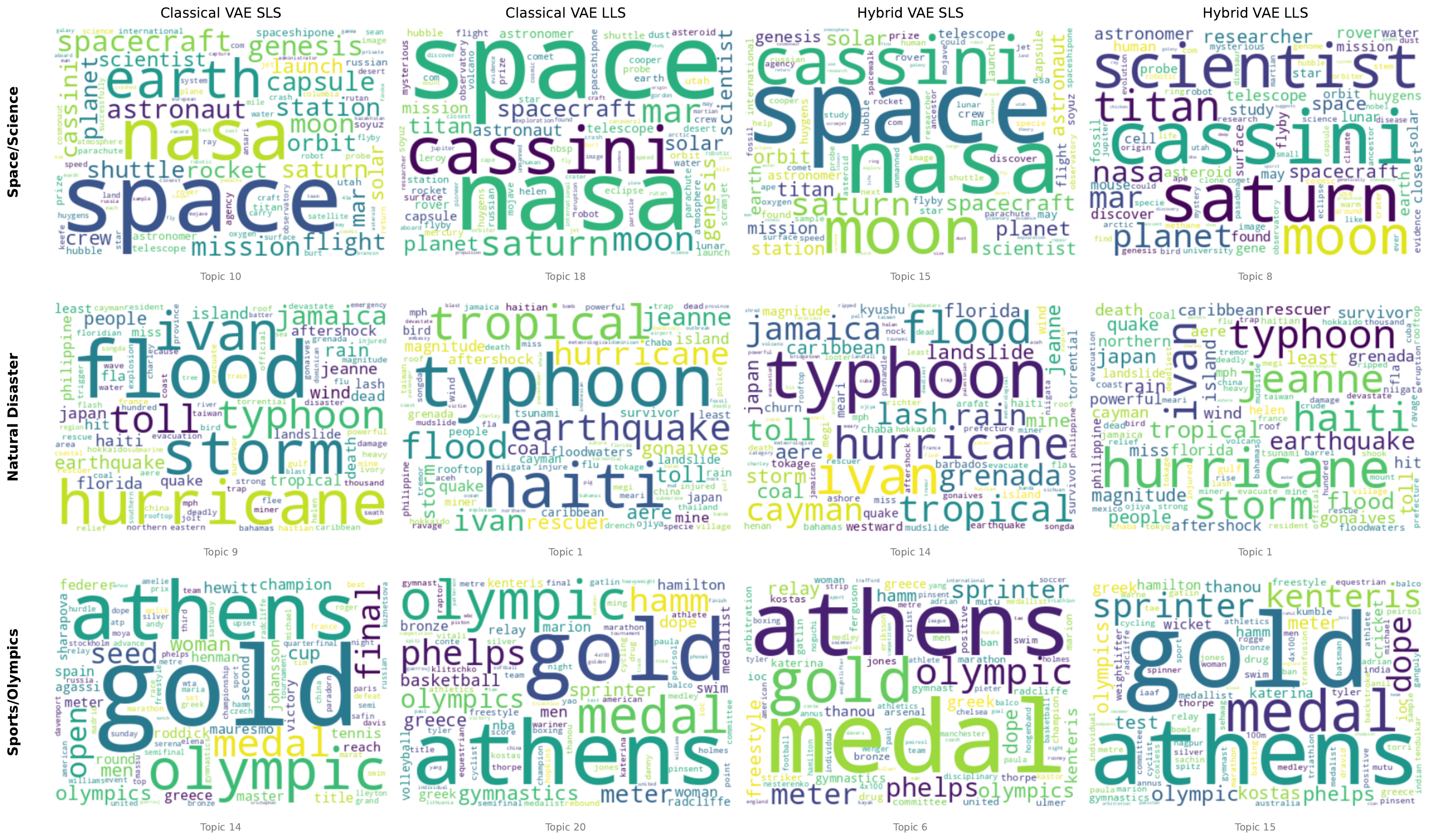}
   \captionof{figure}{Top 100 word clouds for semantic topics across models on AgNews.}
    \label{fig:apd.8}
\end{center}

\begin{center}
   \includegraphics[width=13cm, height=5cm]{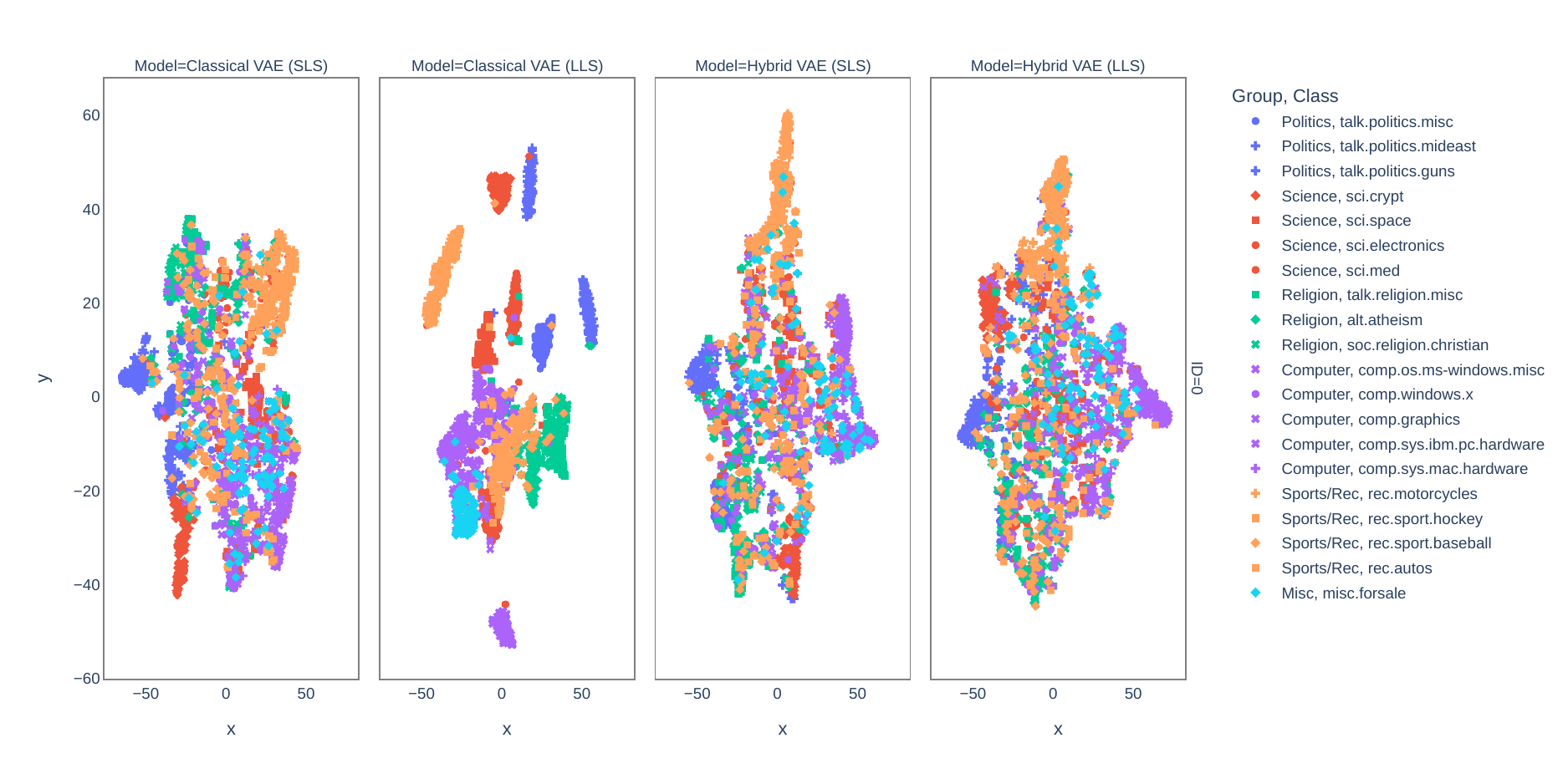}
   \captionof{figure}{2-D t-SNE projection of randomly sampled $z$ from latent spaces under 20 different posterior distributions on 20News.}
    \label{fig:apd.9}
\end{center}

\vspace{.25cm}
\begin{center}
   \includegraphics[width=13cm, height=8cm]{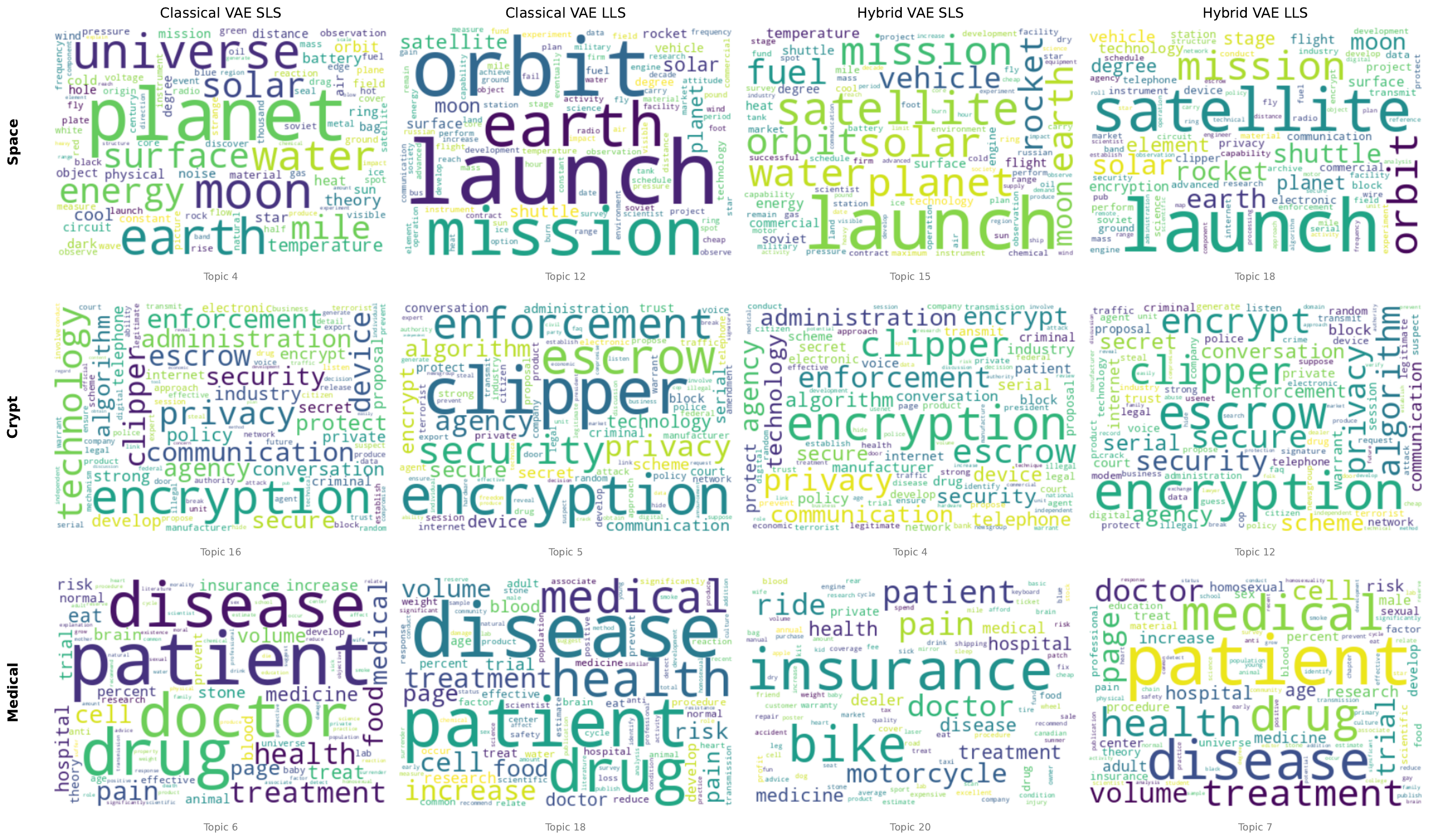}
   \captionof{figure}{Top 100 word clouds for semantic topics across models on 20News.}
    \label{fig:apd.10}
\end{center}

\end{appendices}

\end{document}